\documentclass[runningheads]{llncs}
\usepackage{latexsym}

\usepackage{times}
\usepackage{soul}
\usepackage{url}
\usepackage[hidelinks]{hyperref}
\usepackage[utf8]{inputenc}
\usepackage[small]{caption}
\usepackage{graphicx}
\usepackage{amsmath}
\usepackage{booktabs}
\usepackage{algorithm}
\usepackage{algorithmic}
\usepackage{url,ifthen}
\usepackage{amssymb}

\begin{document}

\title{Generalized Nested Rollout Policy Adaptation}

\author{Tristan Cazenave}

\authorrunning{T. Cazenave}

\institute{LAMSADE, Université Paris-Dauphine, PSL, CNRS, France \email{Tristan.Cazenave@dauphine.psl.eu}}

\maketitle

\begin{abstract}
Nested Rollout Policy Adaptation (NRPA) is a Monte Carlo search algorithm for single player games. In this paper we propose to generalize NRPA with a temperature and a bias and to analyze theoretically the algorithms. The generalized algorithm is named GNRPA. Experiments show it improves on NRPA for different application domains: SameGame and the Traveling Salesman Problem with Time Windows.
\end{abstract}

\section{Introduction}

Monte Carlo Tree Search (MCTS) has been successfully applied to many games and problems \cite{BrownePWLCRTPSC2012}.

Nested Monte Carlo Search (NMCS) \cite{CazenaveIJCAI09} is an algorithm that works well for puzzles and optimization problems. It biases its playouts using lower level playouts. At level zero NMCS adopts a uniform random playout policy. Online learning of playout strategies combined with NMCS has given good results on optimization problems \cite{RimmelEvo11}. Other applications of NMCS include Single Player General Game Playing \cite{Mehat2010}, Cooperative Pathfinding \cite{Bouzy13}, Software testing \cite{PouldingF14}, heuristic Model-Checking \cite{PouldingF15}, the Pancake problem \cite{Bouzy16}, Games \cite{CazenaveSST16} and the RNA inverse folding problem \cite{portela2018unexpectedly}.

Online learning of a playout policy in the context of nested searches has been further developed for puzzles and optimization with Nested Rollout Policy Adaptation (NRPA) \cite{Rosin2011}. NRPA has found new world records in Morpion Solitaire and crosswords puzzles. NRPA has been applied to multiple problems: the Traveling Salesman
with Time Windows (TSPTW) problem \cite{cazenave2012tsptw,edelkamp2013algorithm}, 3D Packing with Object Orientation \cite{edelkamp2014monte}, the physical traveling salesman problem \cite{edelkamp2014solving}, the Multiple Sequence Alignment problem \cite{edelkamp2015monte} or Logistics \cite{edelkamp2016monte}. The principle of NRPA is to adapt the playout policy so as to learn the best sequence of moves found so far at each level.

We now give the outline of the paper. The second section describes NRPA. The third section gives a theoretical analysis of NRPA. The fourth section describes the generalization of NRPA. The fifth section details optimizations of GNRPA. The sixth section gives experimental results for SameGame and TSPTW.

\section{NRPA}

NRPA learns a rollout policy by adapting weights on each action. During the playout phase, action is sampled with a probability proportional to the exponential of the associated weight. The playout algorithm is given in algorithm \ref{PLAYOUT}. The algorithm starts with initializing the sequence of moves that it will play (line 2). Then it performs a loop until it reaches a terminal states (lines 3-6). At each step of the playout it calculates the sum of all the exponentials of the weights of the possible moves (lines 7-10) and chooses a move proportional to its probability given by the softmax function (line 11). Then it plays the chosen move and adds it to the sequence of moves (lines 12-13).

Then, the policy is adapted on the best current sequence found, by increasing the weight of the best actions and decreasing the weights of all the moves proportionally to their probabilities of being played. The Adapt algorithm is given in algorithm \ref{ADAPT}.For all the states of the sequence passed as a parameter it adds $\alpha$ to the weight of the move of the sequence (lines 3-5). Then it reduces all the moves proportionally to $\alpha \times$ the probability of playing the move so as to keep a sum of all probabilities equal to one (lines 6-12).

In NRPA, each nested level takes as input a policy, and returns a sequence. At each step, the algorithm makes a recursive call to the lower level and gets a sequence as a result. It adapt the policy to the best sequence of the level at each step. At level zero it make a playout.

The NRPA algorithm is given in algorithm \ref{NRPA}. At level zero it simply performs a playout (lines 2-3). At greater levels it performs N iterations and for each iteration it calls itself recursively to get a score and a sequence (lines 4-7). If it finds a new best sequence for the level it keeps it as the best sequence (lines 8-11). Then it adapts the policy using the best sequence found so far at the current level (line 12).

NRPA balances exploitation by adapting the probabilities of playing moves toward the best sequence of the level, and exploration by using Gibbs sampling at the lowest level. It is a general algorithm that has proven to work well for many optimization problems.

\begin{algorithm}
\begin{algorithmic}[1]
\STATE{playout ($state$, $policy$)}
\begin{ALC@g}
\STATE{$sequence$ $\leftarrow$ []}
\WHILE{true}
\IF{$state$ is terminal}
\RETURN{(score ($state$), $sequence$)}
\ENDIF
\STATE{$z$ $\leftarrow$ 0.0}
\FOR{$m$ in possible moves for $state$}
\STATE{$z$ $\leftarrow$ $z$ + exp ($policy$ [code($m$)])}
\ENDFOR
\STATE{choose a $move$ with probability $\frac{exp (policy [code(move)])}{z}$}
\STATE{$state$ $\leftarrow$ play ($state$, $move$)}
\STATE{$sequence$ $\leftarrow$ $sequence$ + $move$}
\ENDWHILE
\end{ALC@g}
\end{algorithmic}
\caption{\label{PLAYOUT}The playout algorithm}
\end{algorithm}

\begin{algorithm}
\begin{algorithmic}[1]
\STATE{Adapt ($policy$, $sequence$)}
\begin{ALC@g}
\STATE{$polp \leftarrow$ $policy$}
\STATE{$state \leftarrow$ $root$}
\FOR{$move$ in $sequence$}
\STATE{$polp$ [code($move$)] $\leftarrow$ $polp$ [code($move$)] + $\alpha$}
\STATE{$z$ $\leftarrow$ 0.0}
\FOR{$m$ in possible moves for $state$}
\STATE{$z$ $\leftarrow$ $z$ + exp ($policy$ [code($m$)])}
\ENDFOR
\FOR{$m$ in possible moves for $state$}
\STATE{$polp$ [code($m$)] $\leftarrow$ $polp$ [code($m$)] - $\alpha * \frac{exp (policy [code(m)])}{z}$}
\ENDFOR
\STATE{$state$ $\leftarrow$ play ($state$, $move$)}
\ENDFOR
\STATE{$policy$ $\leftarrow$ $polp$}
\end{ALC@g}
\end{algorithmic}
\caption{\label{ADAPT}The Adapt algorithm}
\end{algorithm}

\begin{algorithm}
\begin{algorithmic}[1]
\STATE{NRPA ($level$, $policy$)}
\begin{ALC@g}
\IF{level == 0}
\RETURN{playout (root, $policy$)}
\ELSE
\STATE{$bestScore$ $\leftarrow$ $-\infty$}
\FOR{N iterations}
\STATE{(result,new) $\leftarrow$ NRPA($level-1$, $policy$)}
\IF{result $\geq$ bestScore}
\STATE{bestScore $\leftarrow$ result}
\STATE{seq $\leftarrow$ new}
\ENDIF
\STATE{policy $\leftarrow$ Adapt (policy, seq)}
\ENDFOR
\RETURN{(bestScore, seq)}
\ENDIF
\end{ALC@g}
\end{algorithmic}
\caption{\label{NRPA}The NRPA algorithm.}
\end{algorithm}

\section{Theoretical Analysis of NRPA}

In NRPA each move is associated to a weight. The goal of the algorithm is to learn these weights so as to produce a playout policy that generates good sequences of moves. At each level of the algorithm the best sequence found so far is memorized. Let ${s_1, ..., s_m}$ be the sequence of states of the best sequence. Let $n_i$ be the number of possible moves in a state $s_i$. Let ${m_{i1}, ...,m_{in_i}}$ be the possible moves in state $s_i$ and $m_{ib}$ be the move of the best sequence in state $s_i$. The goal is to learn to play the move $m_{ib}$ in state $s_i$.

The playouts use Gibbs sampling. Each move $m_{ik}$ is associated to a weight $w_{ik}$. The probability $p_{ik}$ of choosing the move $m_{ik}$ in a playout is the softmax function:

\begin{large}
$$p_{ik} = \frac{e^{w_{ik}}}{\Sigma_j e^{w_{ij}}}$$
\end{large}

The cross-entropy loss for learning to play move $m_{ib}$ is $C_i = -log(p_{ib})$. In order to apply the gradient we calculate the partial derivative of the loss: $\frac{\delta C_i}{\delta p_{ib}} = -\frac{1}{p_{ib}}$. We then calculate the partial derivative of the softmax with respect to the weights:

\begin{large}
$$\frac{\delta p_{ib}}{\delta w_{ij}} = p_{ib} (\delta_{bj} - p_{ij})$$
\end{large}

Where $\delta_{bj} = 1$ if $b = j$ and 0 otherwise. Thus the gradient is:

\begin{large}
$$\nabla w_{ij} = \frac{\delta C_i}{\delta p_{ib}} \frac{\delta p_{ib}}{\delta w_{ij}} = -\frac{1}{p_{ib}} p_{ib} (\delta_{bj} - p_{ij}) = p_{ij} - \delta_{bj}$$
\end{large}

If we use $\alpha$ as a learning rate we update the weights with:

\begin{large}
$$w_{ij} = w_{ij} -\alpha (p_{ij} - \delta_{bj})$$
\end{large}

This is the formula used in the NRPA algorithm to adapt weights.

\section{Generalization of NRPA}

We propose to generalize the NRPA algorithm by generalizing the way the probability is calculated using a temperature $\tau$ and a bias $\beta_{ij}$:

\begin{large}
$$p_{ik} = \frac{e^{\frac{w_{ik}}{\tau} + \beta_{ik}}}{\Sigma_j e^{\frac{w_{ij}}{\tau} + \beta_{ij}}}$$
\end{large}

\subsection{Theoretical Analysis}

The formula for the derivative of $f(x) = \frac{g(x)}{h(x)}$ is:

$$f'(x) = \frac{g'(x)h(x) - h'(x)g(x)}{h^2(x)}$$

So the derivative of $p_{ib}$ relative to $w_{ib}$ is:

$$\frac{\delta p_{ib}}{\delta w_{ib}} = \frac{\frac{1}{\tau}e^{\frac{w_{ib}}{\tau} + \beta_{ib}}\Sigma_j e^{\frac{w_{ij}}{\tau} + \beta_{ij}} - \frac{1}{\tau}e^{\frac{w_{ib}}{\tau} + \beta_{ib}}e^{\frac{w_{ib}}{\tau} + \beta_{ib}}}{(\Sigma_j e^{\frac{w_{ij}}{\tau} + \beta_{ij}})^2}$$

$$\frac{\delta p_{ib}}{\delta w_{ib}} = \frac{1}{\tau}\frac{e^{\frac{w_{ib}}{\tau} + \beta_{ib}}}{\Sigma_j e^{\frac{w_{ij}}{\tau} + \beta_{ij}}} \frac{\Sigma_j e^{\frac{w_{ij}}{\tau} + \beta_{ij}} - e^{\frac{w_{ib}}{\tau} + \beta_{ib}}}{\Sigma_j e^{\frac{w_{ij}}{\tau} + \beta_{ij}}}$$

$$\frac{\delta p_{ib}}{\delta w_{ib}} = \frac{1}{\tau} p_{ib} (1 - p_{ib})$$

The derivative of $p_{ib}$ relative to $w_{ij}$ with $j \neq b$ is:

$$\frac{\delta p_{ib}}{\delta w_{ij}} = -\frac{1}{\tau}\frac{e^{\frac{w_{ij}}{\tau} + \beta_{ij}}e^{\frac{w_{ib}}{\tau} + \beta_{ib}}}{(\Sigma_j e^{\frac{w_{ij}}{\tau} + \beta_{ij}})^2}$$

$$\frac{\delta p_{ib}}{\delta w_{ij}} = -\frac{1}{\tau} p_{ij} p_{ib}$$

We then derive the cross-entropy loss and the softmax to calculate the gradient:

\begin{large}
$$\nabla w_{ij} = \frac{\delta C_i}{\delta p_{ib}} \frac{\delta p_{ib}}{\delta w_{ij}} = -\frac{1}{\tau} \frac{1}{p_{ib}} p_{ib} (\delta_{bj} - p_{ij}) = \frac{p_{ij} - \delta_{bj}}{\tau}$$
\end{large}

If we use $\alpha$ as a learning rate we update the weights with:

\begin{large}
$$w_{ij} = w_{ij} -\alpha \frac{p_{ij} - \delta_{bj}}{\tau}$$
\end{large}

This is a generalization of NRPA since when we set $\tau=1$ and $\beta_{ij}=0$ we get NRPA.

The corresponding algorithms are given in algorithms \ref{PLAYOUTI} and \ref{ADAPTI}. 

\subsection{Equivalence of Algorithms}

Let the weights and probabilities of playing moves be indexed by the iteration of the GNRPA level. Let $w_{nij}$ be the weight $w_{ij}$ at iteration $n$, $p_{nij}$ be the probability of playing move $j$ at step $i$ at iteration $n$, $\delta_{nbj}$ the $\delta_{bj}$ at iteration $n$.

We have:

$$p_{0ij} = \frac{e^{\frac{1}{\tau}w_{0ij}+\beta_{ij}}}{\Sigma_k e^{\frac{1}{\tau}w_{0ik}+\beta_{ik}}}$$

$$w_{1ij} = w_{0ij} - \frac{\alpha}{\tau} (p_{0ij} - \delta_{0bj})$$

$$p_{1ij} = \frac{e^{\frac{1}{\tau}w_{1ij}+\beta_{ij}}}{\Sigma_k e^{\frac{1}{\tau}w_{1ik}+\beta_{ik}}} = \frac{e^{\frac{1}{\tau}w_{0ij}- \frac{\alpha}{\tau^2} (p_{0ij} - \delta_{0bj}) + \beta_{ij}}}{\Sigma_k e^{\frac{1}{\tau}w_{1ik}+\beta_{ik}}}$$

$$w_{2ij} = w_{1ij} - \frac{\alpha}{\tau} (p_{1ij} - \delta_{1bj}) = w_{0ij} - \frac{\alpha}{\tau} (p_{0ij} - \delta_{0bj} + p_{1ij} - \delta_{1bj})$$

By recurrence we get:

$$p_{nij} = \frac{e^{\frac{1}{\tau}w_{nij}+\beta_{ij}}}{\Sigma_k e^{\frac{1}{\tau}w_{nik}+\beta_{ik}}} = \frac{e^{\frac{w_{0ij}}{\tau} - \frac{\alpha}{\tau^2} (\Sigma_k p_{kij} - \delta_{kbj}) + \beta_{ij}}}{\Sigma_k e^{\frac{1}{\tau}w_{nik}+\beta_{ik}}}$$

From this equation we can deduce the equivalence between different algorithms. For example GNRPA$_1$ with $\alpha_1 = (\frac{\tau_1}{\tau_2})^2\alpha_2$ and $\tau_1$ is equivalent to GNRPA$_2$ with $\alpha_2$ and $\tau_2$ provided we set $w_{0ij}$ in GNRPA$_1$ to $\frac{\tau_1}{\tau_2}w_{0ij}$. It means we can always use $\tau = 1$ provided we correspondingly set $\alpha$ and $w_{0ij}$.

Another deduction we can make is we can set $\beta_{ij} = 0$ provided we set $w_{0ij} = w_{0ij} + \tau \times \beta_{ij}$. We can also set $w_{0ij} = 0$ and use only $\beta_{ij}$
which is easier.

The equivalences mean that GNRPA is equivalent to NRPA with the appropriate $\alpha$ and $w_{0ij}$. However it can be more convenient to use $\beta_{ij}$ than to initialize the weights $w_{0ij}$ as we will see for SameGame.



\begin{algorithm}
\begin{algorithmic}[1]
\STATE{playout ($state$, $policy$)}
\begin{ALC@g}
\STATE{$sequence$ $\leftarrow$ []}
\WHILE{true}
\IF{$state$ is terminal}
\RETURN{(score ($state$), $sequence$)}
\ENDIF
\STATE{$z$ $\leftarrow$ 0}
\FOR{$m \in$ possible moves for $state$}
\STATE{$o [m] \leftarrow e^{\frac{policy[code(m)]}{\tau} + \beta(m)}$}
\STATE{$z \leftarrow z + o [m]$}
\ENDFOR
\STATE{choose a $move$ with probability $\frac{o [move]}{z}$}
\STATE{$state \leftarrow play (state, move)$}
\STATE{$sequence \leftarrow sequence + move$}
\ENDWHILE
\end{ALC@g}
\end{algorithmic}
\caption{\label{PLAYOUTI}The generalized playout algorithm}
\end{algorithm}

\begin{algorithm}
\begin{algorithmic}[1]
\STATE{Adapt ($policy$, $sequence$)}
\begin{ALC@g}
\STATE{$polp \leftarrow policy$}
\STATE{$state \leftarrow root$}
\FOR{$b \in sequence$}
\STATE{$z \leftarrow 0$}
\FOR{$m \in$ possible moves for $state$}
\STATE{$o [m] \leftarrow e^{\frac{policy[code(m)]}{\tau} + \beta(m)}$}
\STATE{$z \leftarrow z + o [m]$}
\ENDFOR
\FOR{$m \in$ possible moves for $state$}
\STATE{$polp [code(m)] \leftarrow polp [code(m)] - \frac{\alpha}{\tau}(\frac{o[m]}{z} - \delta_{bm})$}
\ENDFOR
\STATE{$state \leftarrow play (state, b)$}
\ENDFOR
\STATE{$policy \leftarrow polp$}
\end{ALC@g}
\end{algorithmic}
\caption{\label{ADAPTI}The generalized adapt algorithm}
\end{algorithm}

\section{Optimizations of GNRPA}

\subsection{Avoid Calculating Again the Possible Moves}

In problems such as SameGame the computation of the possible moves is costly. It is important in this case to avoid to compute again the possible moves for the best playout in the Adapt function. The possible moves have already been calculated during the playout that found the best sequence. The optimized playout algorithm memorizes in a matrix $code$ the codes of the possible moves during a playout. The cell $code [i] [m]$ contains the code of the possible move of index $m$ at the state number $i$ of the best sequence. The state number 0 is the initial state of the problem. The $index$ array memorizes the index of the code of the best move for each state number, $len(index)$ is the length of the best sequence and $index [i]$ is the index of the best move for state number $i$.

\subsection{Avoid the Copy of the Policy}

Tha Adapt algorithm of NRPA and GNRPA considers the states of the sequence to learn as a batch. The sum of the gradients is calculated for the entire sequence and then applied. The way it is done in NRPA is by copying the policy to a temporary policy, modifying the temporary policy computing the gradient with the unmodified policy, and then copying the modified temporary policy to the policy.

When the number of possible codes is large copying the policy can be costly. We propose to change the Adapt algorithm to avoid to copy twice the policy at each Adapt call. We also use the memorized codes and index so as to avoid calculating again the possible moves of the best sequence.

The way to avoid copying the policy is to make a first loop to compute the probabilities of each move of the best sequence, lines 2-8 of algorithm \ref{OGADAPT}. The matrix $o [i] [m]$ contains the probability for move index $m$ in state number $i$, the array $z[i]$ contains the sum of the probabilities of state number $i$. The second step is to apply the gradient directly to the policy for each state number $i$ and each code, see lines 9-14.


\begin{algorithm}
\begin{algorithmic}[1]
\STATE{Adapt ($policy, code, index$)}
\begin{ALC@g}
\FOR{$i \in [0,len(index)[$}
\STATE{$z [i] \leftarrow 0$}
\FOR{$m \in [0,len(code [i])[$}
\STATE{$o [i] [m] \leftarrow e^{\frac{policy[code[i] [m]]}{\tau} + \beta(m)}$}
\STATE{$z [i] \leftarrow z [i] + o [i] [m]$}
\ENDFOR
\ENDFOR
\FOR{$i \in [0,len(index)[$}
\STATE{$b \leftarrow index [i]$}
\FOR{$m \in [0,len(code [i])[$}
\STATE{$policy [code [i] [m]] \leftarrow policy [code [i] [m]] - \frac{\alpha}{\tau}(\frac{o [i] [m]}{z |i]} - \delta_{bm})$}
\ENDFOR
\ENDFOR
\end{ALC@g}
\end{algorithmic}
\caption{\label{OGADAPT}The optimized generalized adapt algorithm}
\end{algorithm}

\section{Experimental Results}

We now give experimental results for SameGame and TSPTW.

\subsection{SameGame}

The first algorithm we test is the standard NRPA algorithm with codes of the moves using a Zobrist hashing of the cells of the moves \cite{negrevergne2017distributed,edelkamp2016improved,cazenave2016nested}. The selective policy used is to avoid the moves of the dominant color except for moves of size two after move number ten. The codes of the possible moves of the best playout are recorded so as to avoid computing again the possible moves in the Adapt function. It is called NRPA.


Using Zobrist hashing of the moves and biasing the policy with $\beta$ is better than initializing the weights at SameGame since there are too many possible moves and weights. We tried to reduce the possible codes for the moves but it gave worse results. The second algorithm we test is to use Zobrist hashing and the selective policy associated to the bias. It is GNRPA with $\tau=1$ and $\beta_{ij}=min (n - 2 - tabu, 8)$, with $tabu = 1$ if the move is of size 2 and of the tabu color and $tabu = 0$ otherwise. The variable $n$ being the number of cells of the move. The algorithm is called GNRPA.beta.

The third algorithm we test is to use Zobrist hashing, the selective policy, $\beta$ and the optimized Adapt function. The algorithm is called GNRPA.beta.opt.

All algorithms are run 200 times for 655.36 seconds and average scores are recorded each time the search time is doubled.

\begin{figure*}[ht]
    \centering
        \includegraphics[scale=0.70]{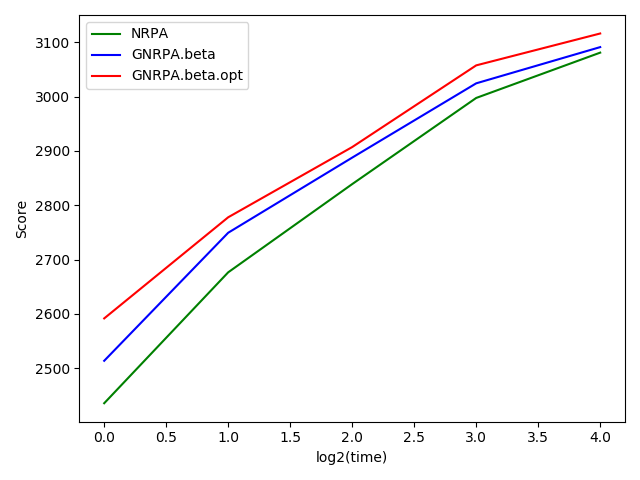}
    \caption{Evolution of the average scores of the three algorithms at SameGame.}
    \label{same}
\end{figure*}

The evolution of the average score of the algorithms is given in figure \ref{same}. We can see that GNRPA.beta is better than NRPA but that for scores close to the current record of the problem the difference is small. GNRPA.beta.opt is the best algorithm as it searches more than GNRPA.beta for the same time. 

Table \ref{tableSame} gives the average scores for the three algorithms associated to the 95\% confidence interval in parenthesis ($2 \times \frac{\sigma}{\sqrt{n}}$).

\begin{table*}[h!]
  \centering
  \caption{Results for the first SameGame problem}
  \label{tableSame}
  \begin{tabular}{lrrrrrrrrrr}
 Time       & ~~~~~~~~~~~~~~~~~~~~NRPA & ~~~~~~~~~~~~GNRPA.beta & ~~~~~~GNRPA.beta.opt) \\
 \\
40.96  & 2435.12 (49.26)  & 2513.35  (53.57)  & 2591.46 (52.50) \\
81.92  & 2676.39 (47.16)  & 2749.33  (47.82)  & 2777.83 (48.05) \\
163.84  & 2838.99 (41.82)  & 2887.78  (39.50)  & 2907.23 (38.45) \\
327.68  & 2997.74 (21.39)  & 3024.68  (18.27)  & 3057.78 (13.52) \\
655.36  & 3081.25 (10.66)  & 3091.44  (10.96)  & 3116.54 (~~7.42) \\
 \end{tabular}

\end{table*}





\subsection{TSPTW}

The Traveling Salesman with Time Windows problem (TSPTW) is a practical problem that has everyday applications. NRPA can be used to efficiently solve practical logistics problems faced by large companies such as EDF \cite{Cazenave2020VRP}.

In NRPA paths with violated constraints can be generated. As presented in \cite{RimmelEvo11} , a new score $Tcost(p)$ of a path $p$ can be defined as follow:
$$
Tcost(p) = cost(p) + 10^6 * \Omega(p),
$$
with, $cost(p)$ the sum of the distances of the path $p$ and $\Omega(p)$ the number of violated constraints. $10^6$ is a constant chosen high enough so that the algorithm first optimizes the constraints.

The problem we use to experiment with the TSPTW problem is the most difficult problem from the set of \cite{potvin1996vehicle}.

In order to initialize $\beta_{ij}$ we normalize the distances  and multiply the result by ten. So $\beta_{ij} = 10 \times \frac{d_{ij}-min}{max-min}$, where $min$ is the smallest possible distance and $max$ the greatest possible one.

All algorithms are run 200 times for 655.36 seconds and average scores are recorded each time the search time is doubled.

\begin{figure*}[ht]
    \centering
        \includegraphics[scale=0.70]{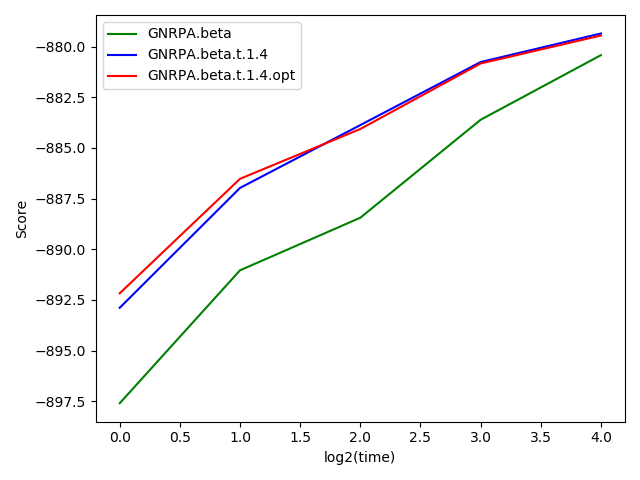}
    \caption{Evolution of the average scores of the three algorithms for TSPTW.}
    \label{tsptw}
\end{figure*}

Figure \ref{tsptw} gives the curves for the three GNRPA algorithms we haves tested with a logarithmic time scale for the x axis.





We could not represent the curve for NRPA in figure \ref{tsptw} since the average values are too low. They are given in table \ref{tableTSPTW}. It is possible to improve much on standard NRPA by initializing the weights with the distances between cities  \cite{edelkamp2013algorithm,Cazenave2020VRP}. However this solution is not practical for all problems as we have seen with SameGame and using a bias $\beta$ is more convenient and general. We also tried initializing the weights with $\beta$ instead of using $\beta$ and we got similar results to the use of $\beta$.

We can see in figure \ref{tsptw} that using a temperature of 1.4 improves on a temperature of 1.0. Using the optimized Adapt function does not improve GNRPA for TSPTW since in the TSPTW problem the policy array and the number of possible moves is very small and copying the policy is fast.

The curve of the best algorithm is asymptotic toward the best value found by all algorithms. It reaches better scores faster.

Table \ref{tableTSPTW} gives the average values for NRPA and the three GNRPA algorithms we have tested. As there is a penalty of 1 000 000 for each constraint violation, NRPA has very low scores compared to GNRPA. This is why NRPA is not depicted in figure \ref{tsptw}. For a search time of 655.36 seconds and not taking into account the constraints, NRPA usually reaches tour scores between -900 and -930. Much worse than GNRPA. We can observe that using a temparature is beneficial until we use 655.36 seconds and approach the asymptotic score when both algorithms have similar scores. The numbers in parenthesis in the table are the 95\% confidence interval ($2 \times \frac{\sigma}{\sqrt{n}}$).

\begin{table*}[h!]
  \centering
  \caption{Results for the TSPTW rc204.1 problem}
  \label{tableTSPTW}
  \begin{tabular}{rrrrrrrrrrr}
 Time       &  NRPA & GNRPA.beta & GNRPA.beta.t.1.4 & GNRPA.beta.t.1.4.opt \\
 \\
40.96  & -3745986.46 (245766.53 )  & -897.60  (1.32 )  & -892.89 (0.96 )  & -892.17 (1.04 ) \\
81.92  & -1750959.11 (243210.68 )  & -891.04  (1.05 )  & -886.97 (0.87 )  & -886.52 (0.83 ) \\
163.84  & -1030946.86 (212092.35 )  & -888.44  (0.98 )  & -883.87 (0.71 )  & -884.07 (0.70 ) \\
327.68  & -285933.63 (108975.99 )  & -883.61  (0.63 )  & -880.76 (0.40 )  & -880.83 (0.32 ) \\
655.36  & -45918.97 (38203.97 )  & -880.42  (0.30 )  & -879.35 (0.16 )  & -879.45 (0.17 ) 
 \end{tabular}
\end{table*}



\section{Conclusion}

We presented a theoretical analysis and a generalization of NRPA named GNRPA. GNRPA uses a temperature $\tau$ and a bias $\beta$. 

We have theoretically shown that using a bias is equivalent to initializing the weights. For SameGame initializing the weights can be difficult if we initialize all the weights at the start of the program since there are too many possible weights, whereas using a bias $\beta$ is easier and improves search at SameGame. A lazy initialization of the weights would also be possible in this case and would solve the weight initialization problem for SameGame. For some other problems the bias could be more specific than the code of the move, i.e. a move could be associated to different bias depending on the state. In this case different bias could be used in different states for the same move which would not be possible with weight initialization. 

We have also theoretically shown that the learning rate and the temperature can replace each other. Tuning the temperature and using a bias has been very beneficial for the TSPTW.

The remaining work is to apply the algorithm to other domains and to improve the way to design formulas for the bias $\beta$.

\bibliographystyle{splncs04}
\bibliography{main}
\end{document}